\documentclass{Interspeech2024}
\usepackage{bbm}
\usepackage{subfig}
\usepackage{multirow}
\usepackage{booktabs}
\usepackage{cite}
\usepackage{bm}

\interspeechcameraready

\title{A Parameter-efficient Language Extension Framework for Multilingual ASR}

\name[affiliation={1}]{Wei}{Liu$^{*,}$}
\name[affiliation={2}]{Jingyong}{Hou}
\name[affiliation={2}]{Dong}{Yang}
\name[affiliation={2}]{Muyong}{Cao}
\name[affiliation={1}]{Tan}{Lee}
\address{
  $^1$ Department of Electronic Engineering, The Chinese University of Hong Kong \\
  $^2$ GVoice, Tencent
  }
\email{
\thanks{$^{*}$ This work was done during an internship at Tencent.}
louislau\_1129@link.cuhk.edu.hk, \{jingyonghou,daviddyang,locwellcao\}@tencent.com, tanlee@cuhk.edu.hk}


\begin{document}

\maketitle

\begin{abstract}
    
Covering all languages with a multilingual speech recognition model (MASR) is very difficult. Performing language extension on top of an existing MASR is a desirable choice. 
In this study, the MASR continual learning problem is probabilistically decomposed into language identity prediction (LP) and cross-lingual adaptation (XLA) sub-problems. Based on this, we propose an architecture-based framework for language extension that can fundamentally solve catastrophic forgetting, debudded as PELE. PELE is designed to be parameter-efficient, incrementally incorporating an add-on module to adapt to a new language. Specifically, different parameter-efficient fine-tuning (PEFT) modules and their variants are explored as potential candidates to perform XLA. Experiments are carried out on 5 new languages with a wide range of low-resourced data sizes. The best-performing PEFT candidate can achieve satisfactory performance across all languages and demonstrates superiority in three of five languages over the continual joint learning setting. Notably, PEFT methods focusing on weight parameters or input features are revealed to be limited in performance, showing significantly inferior extension capabilities compared to inserting a lightweight module in between layers such as an Adapter.

\end{abstract}

\vspace{-5pt}
\section{Introduction}
Multilingual automatic speech recognition (MASR) refers to a process in which a single model can transcribe speech in multiple languages~\cite{DBLP:conf/interspeech/PratapSTHLSC20, li2021scaling, li2022massively, DBLP:conf/lrec/YadavS22}. It has been shown that tens and even hundreds of languages can be supported in one unified model, given the success of Whisper~\cite{radford2023robust} and USM~\cite{zhang2023google}. 
Nevertheless, in practice, there are always new languages not currently covered by the model.
After collecting the labeled data of new target languages, instead of training a monolingual ASR, extending the original MASR to new languages is an intuitively better choice to leverage the already encoded multilingual knowledge.

Extending an existing MASR to new languages lies in the field of continual learning (CL)~\cite{awasthi2019continual, de2021continual, kirkpatrick2017overcoming, li2017learning, rolnick2019experience, DBLP:conf/iclr/ChaudhryRRE19, rusu2016progressive, mallya2018piggyback}. CL aims to learn a sequence of tasks incrementally. The major challenge in CL is catastrophic forgetting (CF). The tuned model weights after learning a new task would hinder the performance of the preceding tasks. CL has been investigated in computer vision~\cite{wortsman2020supermasks, qu2021recent} and natural language processing (NLP)~\cite{DBLP:conf/coling/BiesialskaBC20, DBLP:conf/iclr/RazdaibiedinaMH23, wang2023rehearsal}, mostly on classification tasks. It is far less explored in the scope of ASR~\cite{vander2022continual, DBLP:conf/icassp/Eeckth23}, especially for multilingual ASR as a class incremental learning (CIL) setting.  Unlike domain incremental learning (DIL) where the input data distribution changes and the output classes remain the same, CIL introduces new class labels, i.e., including new language tokens into vocabulary. Thus supporting transcription in a new language is a more difficult CL problem, compared with adapting to new domains \cite{vander2022continual} or learning new tasks with much similar inputs \cite{wortsman2020supermasks, DBLP:conf/iclr/RazdaibiedinaMH23, wang2023rehearsal}.



Previous studies categorize the approaches to CL into three-folds: \textit{regularization}-based \cite{kirkpatrick2017overcoming, li2017learning}, \textit{rehearsal}-based \cite{rolnick2019experience, DBLP:conf/iclr/ChaudhryRRE19} and \textit{architecture}-based \cite{rusu2016progressive, mallya2018piggyback, wortsman2020supermasks, DBLP:conf/iclr/RazdaibiedinaMH23, wang2023rehearsal}. These methods themselves are usually task-agnostic. Considering the aforementioned challenge, their effectiveness in the ASR language extension remains to be verified. 
Della~\cite{della2023cl} recently established a CL benchmark for MASR and explored different CL methods. It was found that \textit{rehearsal}-based and \textit{architecture}-based are more effective than \textit{regularization}-based methods. In~\cite{della2023cl}, 10 hours of adaptation data were used for a new language. 
The averaged word error rates of the extended MASR systems are above $40\%$ (mostly around $65\%$). 
The performance is below the level for practical use. Speech data of ten hours are considered extremely limited in an industrial setting even from the low-resource perspective. 
It remains unclear whether satisfactory performance can be attained when dealing with a more practical data size or if these methods are simply limited in performance. 

In this paper, we investigate an architecture-based CL framework to facilitate language extension on a MASR model. 
The existing MASR model acts as a base model, incrementally integrating an add-on module for adapting to a new language. 
Without the need to access the original training data of previous languages, the parameters of the base model are frozen to fundamentally avoid CF. Experiments on language extension are carried out with 5 new languages, for which the amount of data ranges from $22$ to $284$ hours. To reduce the storage overhead in deployment, the proposed framework \textbf{PELE} (abbreviating \textbf{P}arameter \textbf{E}fficient \textbf{L}anguage \textbf{E}xtension) is designed to be parameter-efficient and scalable in terms of the number of extended languages. While~\cite{della2023cl} did not specifically consider this scalability. It is shown that, with around 10M parameters per language, the best add-on module candidate in PELE can exhibit superior performance on three of five languages compared to the continual joint training setting. 
To conclude, the major contributions of this study are summarized below:
\begin{itemize}[leftmargin=*]
    \item We propose a theoretically inspired \cite{kim2022theoretical} CL framework for language extension in multilingual ASR, in which the original MASR continual learning problem is probabilistically decomposed into two sub-problems, language identity prediction (LP) and cross-lingual adaptation (XLA).
    \item We propose utilizing the parameter-efficient fine-tuning (PEFT) methods and their variants to adapt to new languages in PELE to achieve a trade-off between the method's scalability and representation capacity.   
    
    \item Our proposed PELE exhibits significantly better performances compared to several competitive baselines, with a very limited increase in the number of parameters. The experiments are conducted using a wider range of low-resourced data sizes, making the conclusions more instructive.
    
    
\end{itemize}

\vspace{-5pt}
\section{PELE} 


\subsection{CL-MASR Problem Decomposition}
Denote an input speech feature sequence as $\mathbf{X}$ and its corresponding transcription as $\mathbf{Y}$, speech recognition can be formulated as $p(\mathbf{Y}|\mathbf{X})$. 
In the multilingual scenario, a latent variable $l$ is introduced to represent spoken language identity. According to the Bayesian formula, MASR can be formulated as:
\begin{equation}
\label{eq:decomp_total}
    p(\mathbf{Y}|\mathbf{X}) = \sum_{l\in L}{p(\mathbf{Y},l|\mathbf{X})}= \sum_{l \in L}p(l|\mathbf{X})p(\mathbf{Y}|\mathbf{X}, l),
\end{equation}
where $L$ is the set of all supported languages. 
Let's consider a new target language $l_i$ that we aim to extend. We make an assumption, denoted as \textbf{Asmp. 1}, that ``$p(\mathbf{Y},l_j|\mathbf{X}) = 0$ for all pairs $(\mathbf{X}, \mathbf{Y})$ that belong to language $l_i$, if $l_j \neq l_i$". Under this assumption, the conditional probability $p(\mathbf{Y}|\mathbf{X}) = p(\mathbf{Y},l_i|\mathbf{X})$. 
This assumption simplifies the interaction of information, such as language-specific modules and vocabulary, between the new language $l_i$ and other languages. 
Thus when we want to support a new language $l_i$ based on an existing MASR model, the negative log-likelihood $-\log{p(\mathbf{Y}|\mathbf{X})}$ serves as the continual learning loss $\mathcal{L}_{CL}$ to be minimized. Based on Eq. \ref{eq:decomp_total} and \textbf{Asmp. 1}, $\mathcal{L}_{CL}$ can be decomposed into two separate terms:
\begin{equation}
    \mathcal{L}_{CL} = \mathcal{L}_{LP} + \mathcal{L}_{XLA} =  -logp(l_i|\mathbf{X}) -logp(\mathbf{Y}|\mathbf{X}, l_i).
\end{equation}
Here the first term $\mathcal{L}_{LP}$ signifies the loss for language identity prediction (LP) and the second term $\mathcal{L}_{XLA}$ represents the loss of cross-lingual adaptation (XLA) for the language $l_i$. The overall loss $\mathcal{L}_{CL}$ is upper-bounded by $\mathcal{L}_{LP}$ and $\mathcal{L}_{XLA}$. 
Therefore, improving either of these two terms would contribute to the performance of newly supported languages. With this in mind, we break down the CL-MASR problem into two sub-problems for more effective analysis and modeling.
They are (i) LP, to identify the spoken language of input speech;  
and (ii) XLA, to adapt the original MASR model for the new language. 
\subsection{CL Modification to MASR}
In this section, we demonstrate how to enable CL on an existing MASR model by explicitly incorporating LP and XLA. In this study, the existing MASR model, before CL modifications, adopts a typical hybrid CTC-attention architecture~\cite{watanabe2017hybrid}, consisting of an encoder, a decoder, and a CTC layer. The encoder first converts the speech feature $\mathbf{X}$ to hidden representation $\mathbf{H}$. Then  $\mathbf{H}$ is forwarded to two output branches, the CTC layer and decoder, for obtaining the final transcription $\mathbf{Y}$.


LP is a classification task that can be trained separately. To seamlessly incorporate this function, leveraging the existing encoder to classify LID is expected.  Without any changes to the base model, the $n$-th layer output $\mathbf{H}^{(n)}$ is extracted as features to discriminate languages. 
We attempt two kinds of approaches: (1) neural network-based (MLP): A three-layer MLP takes $\mathbf{H}^{(n)}$ as input for LID classification. (2) Gaussian discriminative analysis (GDA) \cite{wang2023rehearsal}: a non-parametric method that models each language class as a Gaussian distribution with the estimated mean and covariance from $\mathbf{H}^{(n)}$s. The classification is then performed by comparing the test sample's $\mathbf{H}^{(n)}$ with all languages' Gaussians and choosing the nearest one.

XLA is the core function of extending a new language for MASR. 
Architecture-based CL suggests incrementally assigning a specific module for each language. 
On the other hand, many parameter-efficient fine-tuning (PEFT) modules exhibit powerful learning capacity on par with full fine-tuning on a wide range of NLP 
tasks~\cite{DBLP:conf/iclr/HuSWALWWC22, houlsby2019parameter, DBLP:conf/emnlp/LesterAC21}. 
The scalability of PEFT makes it easy to extend more languages. Thus
PEFT can serve as a potential candidate to perform high-quality XLA.  
PEFT essentially injects a small set of trainable parameters into the original model in various forms, while keeping the original base model frozen. 
According to the different injection positions, they can be divided into three categories: (1) \textit{parameter composition}, (2) \textit{function composition}, and (3) \textit{input composition}.


\vspace{-5pt}
\subsection{PELE Formulation}
To coordinately enable LP and XLA together, a general framework PELE is thus proposed. 
The multilingual information interaction among the add-on PEFT modules of XLA and the base model can be naturally included in this framework without considering \textbf{Asmp. 1}. According to Eq. \ref{eq:decomp_total}, our PELE can be written in a module-level formulation as follows: \vspace{-5pt}
\begin{equation}
\vspace{-3pt}
\label{eq:pele}
    \mathbf{m}^{*} = \sum_{l=0}^{L^{'}} \alpha_l \mathbf{m}_l,
\end{equation}
where $\mathbf{m}_l$ indicates a specific PEFT module to adapt on the language $l$ and $\alpha_l$ is the corresponding weight coefficient. $L^{'}$ denotes the number of new extended languages. 
$L^{'} + 1$ add-on modules are placed in parallel, similar to the structure of mixture-of-expert (MoE)~\cite{liu2023moelora}, for the extension of multiple languages. $l=0$ represents the base languages that the original MASR model supported. Note that $\mathbf{m}_0$ serves as a dummy forward, lacking trainable parameters but being a module equivalent in form to $\{\mathbf{m}_l\}_{l=1}^{L^{'}}$. In this manner, it can seamlessly revert to the original MASR without any adaptation effect when $\bm{\alpha} = [1,0,...0]$ (we denote the vector $\bm{\alpha} = [\alpha_0, \alpha_1, ..., \alpha_{L^{'}}]$). 

By comparing Eq. \ref{eq:pele} with Eq. \ref{eq:decomp_total}, 
$\mathbf{m}_l$ can be regarded as a module-level realization of $p(\mathbf{Y}|\mathbf{X},l)$ for language adaptation, and the coefficient $\alpha_i$ is viewed to approximate $p(l|\mathbf{X})$. It is worth noting that the coefficient vector $\bm{\alpha}$ is not necessarily to be derived from LP posterior. Other choices like the ground-truth one-hot vector (pre-know the language identity) or entirely the learnable vector can also be attempted.  With a certain $\bm{\alpha}$, the output $\mathbf{m}^{*}$ represents the final module-level adaptation.

Different PEFT choices result in different formulations of $\mathbf{m}$. Specifically, (1) in \textit{parameter composition} PEFT, $\mathbf{m}$ is usually the injected parameter matrix, such as the incremental update $\Delta \mathbf{W}$ of LoRA \cite{DBLP:conf/iclr/HuSWALWWC22} or the mask weight of Mask tuning \cite{fu2022losses, yu2023master}. (2) In \textit{function composition} PEFT, $\mathbf{m}$ is exactly the inserted lightweight module in-between layers, such as $Adapter(\mathbf{H}^{n})$ \cite{houlsby2019parameter}. (3) In \textit{input composition} PEFT, $\mathbf{m}$ can be the virtual token sequence of prompt tuning \cite{DBLP:conf/emnlp/LesterAC21}. In general,  Eq. \ref{eq:pele} is flexible to be implemented and incorporated into any specified layer/sub-layer of the base model for adaptation.


\section{Experimental Setup}
\subsection{Base Model \& Dataset}
Consider a given MASR base model that adopts a hybrid CTC-attention architecture~\cite{watanabe2017hybrid}. It has 12 conformer layers~\cite{DBLP:conf/interspeech/GulatiQCPZYHWZW20}  as encoder and 6 transformer layers~\cite{vaswani2017attention} as decoder. The attention dimension is 512. The based model is trained on a dataset which is a mixture of 10 languages with 4714.1 hours in total. The language includes English (en), French (fr), Spanish (es), Chinese (zh), Italian (it), Russian (ru), Portuguese (pt), Turkish (tr), Dutch (nl), and Tatar (tt) from the Common Voice 13.0~\cite{DBLP:conf/lrec/ArdilaBDKMHMSTW20}. In our CL MASR experiments, we select 5  `never-seen' languages in the Common Voice, to perform language extension. They are German (de, 284.3h), Polish (pl, 131.7h), Welsh (cy, 101.5h),  Japanese (ja, 55.5h), and Czech (cs, 22.4h). The number in brackets denotes the training data size in terms of hours.  

\subsection{MASR Continual Learning}
Extending the tokens for unseen languages is a crucial step in CL-MASR.
To construct an expanded token embedding matrix, the embeddings of new tokens are usually randomly initialized and then concatenated with the original matrix.
In our experiments, for simplicity, the Whisper~\cite{radford2023robust} tokenizer is used to generate the output vocabulary at once. In the base model, only the tokens of those 10 seen languages were explicitly trained. 

In the CL training procedure, Adam~\cite{DBLP:journals/corr/KingmaB14} optimizer with an initial learning rate of $1e-3$ is used. The warmup steps are $2000$. The batch\_size is 12 with an $accum\_grad$ of 8. Eight V100 GPUs are used for DDP training. If not specified otherwise, the number of training epochs is $50$ and the CTC greedy decoding results are reported. For the evaluation metric, Word or character error rate (WER/CER) is adopted to measure the performance of different extended MASR systems. 

\subsubsection{Baselines}
\textit{\textbf{Mono}}: A monolingual ASR model is trained for each language. The attention dimension is 256.
\textit{\textbf{Raw}}: Directly use the base model for recognition. \textit{\textbf{FullFT}}: Fully fine-tune the base model with the data of 5  new languages. \textit{\textbf{CJT}}: Continually joint training with the mixture of 5  new languages and the original 10 languages on top of the base model. \textit{\textbf{ER}}: Experience Replay is a typical rehearsal-based CL method, showing the best CL-MASR performance in~\cite{della2023cl}. To alleviate CF, a certain amount of historical data per language is allowed to be stored for mixing with the current new language data in fine-tuning.
\subsubsection{PELE}
In LP, at most 100k utterances per language are randomly selected to perform MLP or GDA based LID prediction. For the original 10 base languages, we assume the LID label can be obtained.
By observing LP results, the 6th encoder layer output $\mathbf{H}^{6}$ is selected as the feature to identify LID. For XLA, the adaptation modification is thus performed starting from the 7th encoder layer. 
Different PEFT modules are explored in the PELE framework to perform cross-lingual adaptation. We illustrate them as follows: (a) For \textit{parameter composition} type,  
\textit{\textbf{BitFit}} \cite{DBLP:conf/acl/ZakenGR22}: Only the bias parameters of the base model are updated. 
\textit{\textbf{LoRA}} \cite{DBLP:conf/iclr/HuSWALWWC22}: The incremental update $\Delta \mathbf{W}=\mathbf{B}\mathbf{A}$ is assumed to be low-rank decomposed, where $\mathbf{B} \in \mathbb{R}^{d_2 \times r}$ and $\mathbf{A} \in \mathbb{R}^{r \times d_1}$. In our experiments, another low-rank decomposed matrix $\mathbf{W}_{s}$ is introduced to adjust the original weight matrix $\mathbf{W}_0$, i.e., $\mathbf{W} = \mathbf{W}_0 \odot \mathbf{W}_{s} + \Delta \mathbf{W}$, where $\odot$ denotes the element-wise product.  We applied this improved LoRA version with $r=32$ on all weight matrices of attention layers and feedforward layers. \textit{\textbf{LoRA*}}: Increase the rank of LoRA to 128 for the output matrix of attention layers and the down projection matrix of feedforward layers. \textit{\textbf{Mask}} \cite{yu2023master}: Instead of adjusting the weight values, mask tuning opts to change the weight connection by multiplying $\mathbf{W}_0$ with a binary mask $\mathbf{B} \in \{0,1\}^{d_2 \times d_1}$. $\mathbf{B}$ is derived by a learnable matrix with a threshold. The weight matrices this method works on are the same as LoRA, and the masking threshold is set as $0.05$.  \textit{\textbf{MaskLoRA*}}: We propose the combination of Mask and LoRA*, i.e.,  $\mathbf{W} = (\mathbf{W}_0 \odot \mathbf{W}_{s} + \Delta \mathbf{W}) \odot \mathbf{B}$.  (b) For \textit{function composition} type, \textit{\textbf{Adapter}} \cite{houlsby2019parameter}: Inserted after the self-attention layer and feedforward layer sequentially. The bottleneck dimension is $256$. (c) For \textit{input composition} type, \textit{\textbf{Prompt}} \cite{DBLP:conf/emnlp/LesterAC21}: A sequence of 20 learnable token embeddings are prepended to input speech feature sequences to serve as new inputs.  

It is noted that to achieve the overall parameter efficiency, the vocabulary layer with large parameters is suggested to have a low-rank update. For this purpose, Eq. \ref{eq:pele} is implemented on the vocabulary layer, where $\mathbf{m}_l$ adopts LoRA ($r=32$) to apply on the token embedding matrix for the language $l$.

\begin{figure}[t!]
  \centering
  \resizebox{0.97\linewidth}{!}{
  \subfloat[LP on the original 10 languages ]
   {\includegraphics[width=0.247\textwidth]{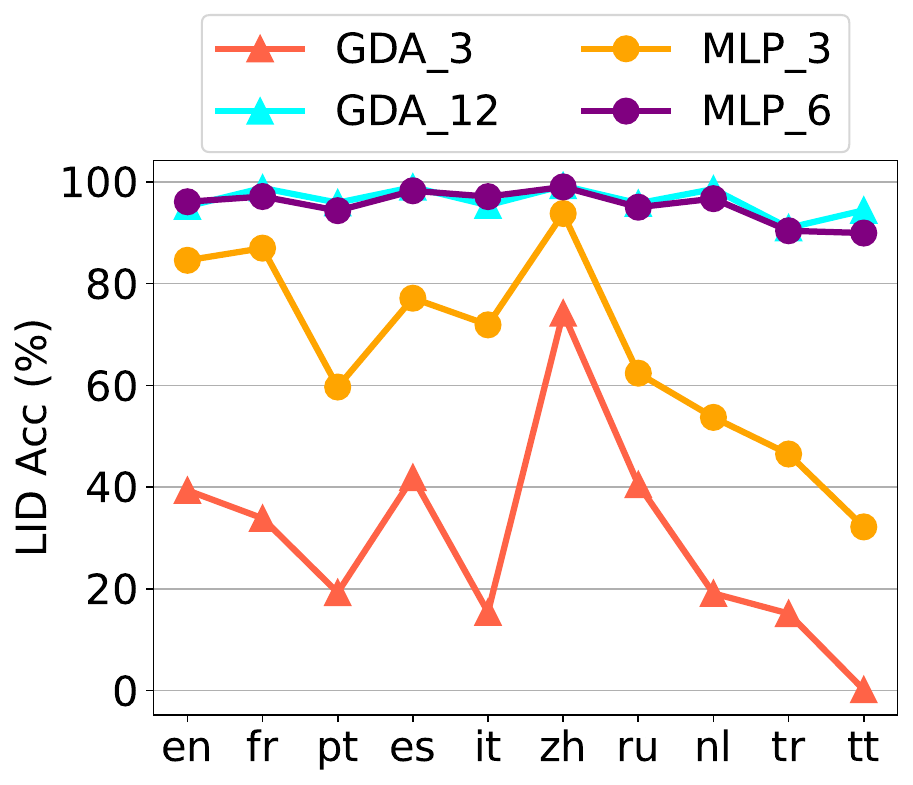}}
  \subfloat
    [LP on the new 5  languages]
   {\includegraphics[width=0.247\textwidth]{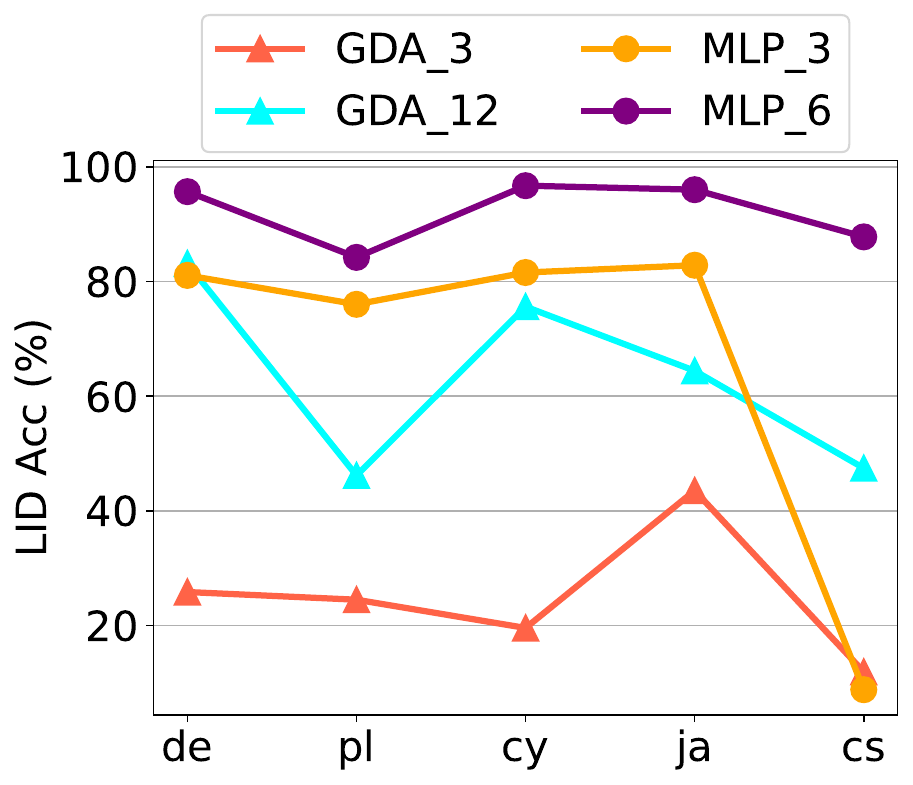}}
   }
    \resizebox{0.97\linewidth}{!}{
  \subfloat
    [XLA on ja with LoRA-vocab]
   {\includegraphics[width=0.255\textwidth]{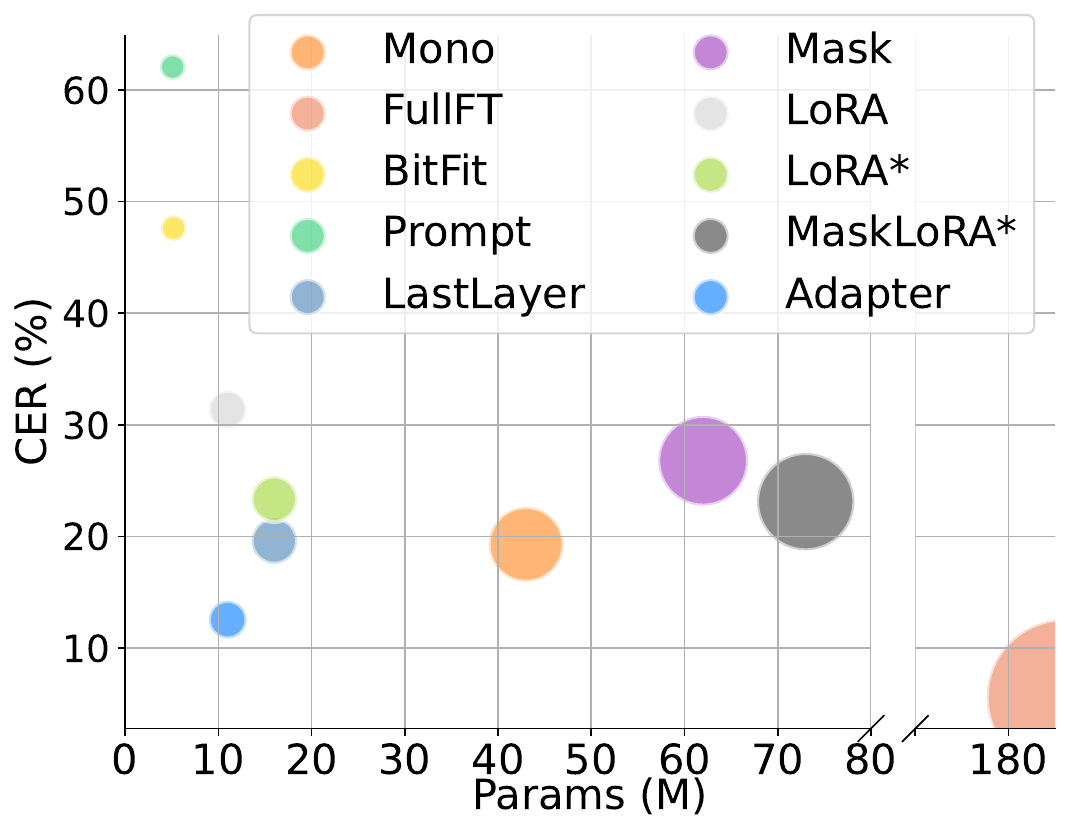}}
  \subfloat
    [XLA on ja with Full-vocab]
   {\includegraphics[width=0.245\textwidth]{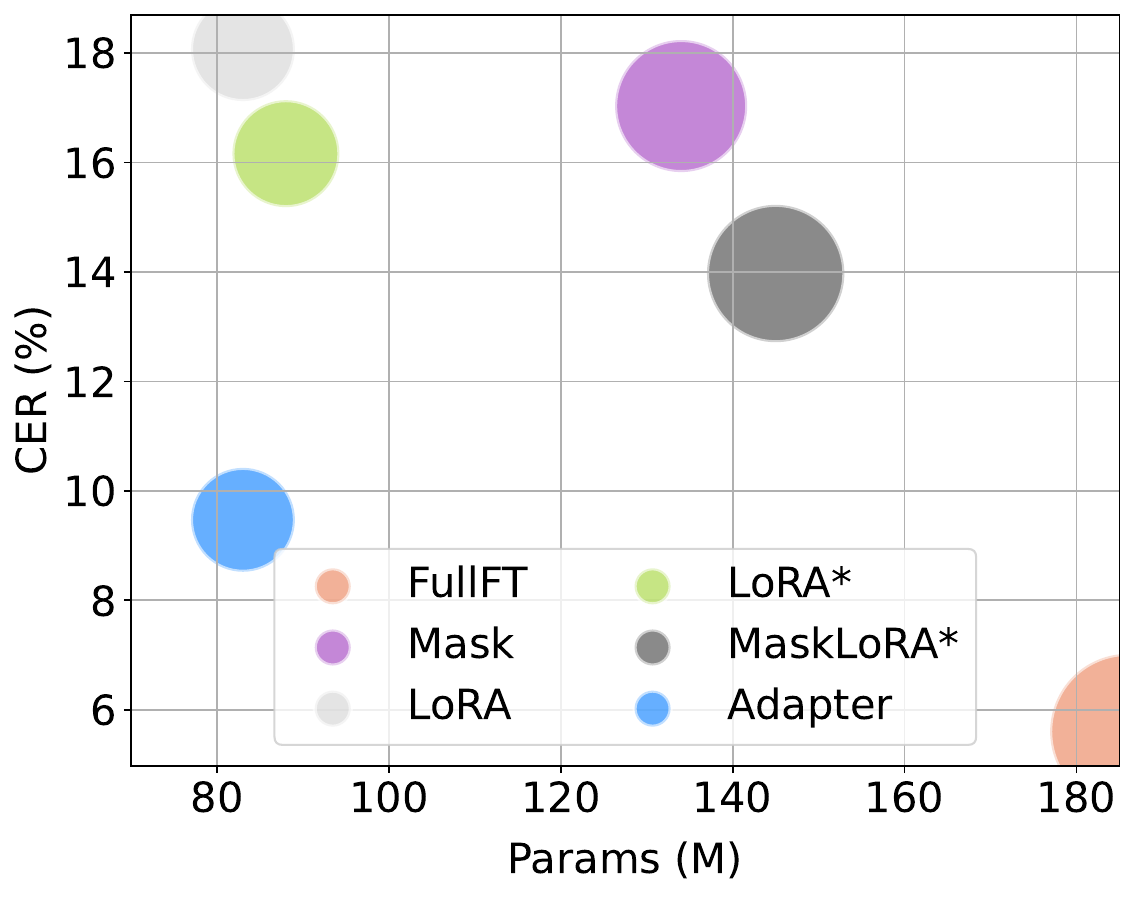}}
   }
  
  \caption{First row: The accuracy of different LID prediction methods. (a) on the original 10 languages; (b) on the new 5 languages. Second row: The CER results of different PEFT methods and baselines when performing XLA on the unseen language Japanese (ja). (c) the vocabulary layer is low-rank updated; (d) the vocabulary layer is fully updated.}
\label{fig:lp_xla}
\vspace{-5mm}
\end{figure}

\section{Results and Analysis}
\subsection{Results of the Decomposed Sub-problems}
\subsubsection{Language Prediction (LP)}
Fig. \ref{fig:lp_xla} (a) and (b) plot the LID prediction accuracy on the original 10 languages and the new 5 languages, respectively. GDA and MLP are two kinds of language classification methods; the former is a non-parametric approach. The suffix number represents which encoder layer output is utilized as the feature. In the original 10 languages, GDA\_12 approaches 100\% accuracy, while the results of GDA\_3 are mostly under 50\% accuracy.  Similarly, MLP\_6 largely outperforms MLP\_3. It implies that the higher encoder layer has richer information to discriminate languages.  In the new 5 languages, the performance of GDA\_12 exhibits significant degradation. It is hypothesized that the layer representation of the base model is hard to generalize to unseen languages without any training. In contrast, MLP\_6 maintains a reasonable prediction accuracy. Not using MLP\_12 is because of the trade-off that several higher encoder layers have to be kept to perform adaptation modification.

\subsubsection{Cross-lingual Adaptation (XLA)}
\label{subsubsec: XLA}
In XLA, a never-seen language, Japanese (ja), is utilized as the target language for adaptation. Fig. \ref{fig:lp_xla} (c) shows the XLA results of various PEFT methods and some baselines by training $100$ epochs. The y-axis denotes the CER 
performance, the lower, the better. The x-axis denotes the number of trainable parameters to measure the parameter efficiency. The circles with different colors represent different adaptation methods. The scalability of the method is better if the corresponding circle is smaller as fewer parameters are required to adapt to a new language.

\textit{Prompt} and \textit{BitFit} exhibit the two smallest circles, showing the worst adaptation performance. We argue that their representation capacity is quite limited to realize the MASR language extension. \textit{LoRA} gives a CER of around 30\% and the enhanced \textit{LoRA*} further decrease the CER to 23.3\%, demonstrating the importance of the PEFT's capacity. For the method of mask tuning, it can be seen that \textit{Mask} gives a CER result in between \textit{LoRA} and \textit{LoRA*}. At the cost of more trainable parameters, \textit{MaskLoRA*} surpasses \textit{Mask} and \textit{LoRA*} by marrying the strengths of both sides. The method \textit{LastLayer} denotes updating the last encoder and decoder layer of the base model, achieving a strong result. The blue circle located in the bottom left corner, in particular, draws our attention. This \textit{Adapter} method can attain 12.5\% CER with 11M parameters. Its performance only left behind \textit{FullFT}, while significantly increasing the adaptation scalability. In Fig. \ref{fig:lp_xla} (d), instead of low-rank updating the vocabulary (LoRA-vocab), several potential PEFT methods are selected to perform adaptation with the vocabulary fully updated. We observe that \textit{Adapter} still keeps the obvious CER advantage and shows the least performance decrease when comparing LoRA-vocab to Full-vocab. 

The overall XLA results suggest that different from the previous CL problems studied in NLP, for the ASR language extension, \textit{parameter} and \textit{input composition} type of PEFT methods, e.g., \textit{LoRA}~\cite{DBLP:conf/iclr/HuSWALWWC22}, \textit{Mask}~\cite{yu2023master}, \textit{Prompt}~\cite{DBLP:conf/emnlp/LesterAC21},  are performance-limited, showing significantly less representation capacity than \textit{Adapter}~\cite{houlsby2019parameter}, a \textit{function composition} PEFT method.

\begin{table}[t!]
\centering
\caption{The WER (\%) performance comparison of MASR continual learning between baselines and the proposed PELE. $base10$ represents the average recognition performance on the original 10 languages that the base model is trained with. $Avg$ represents the averaged WER of all 15 languages. $Inc. ~Params$ denotes the averaged parameters that increased by supporting a new language. ER (1k) denotes 1k utterances per language are cached for experience replay, and ER (10k) is similar. In the $\bm{\alpha}$ column, $LP~post$ means $\bm{\alpha}$ adopts the LID prediction posterior, while $LP~ohot$ denotes the one-hot vector derived by $LP~post$ is used. $GT~ohot$ means adopting the ground-truth one-hot vector and $GT~learn$ represents the learnable vector parameters that first initialized as $GT~ohot$.        }

\resizebox{1.01\linewidth}{!}{
\begin{tabular}{cc|cccccccc}
\toprule
\multicolumn{2}{c|}{\multirow{2}{*}{Methods}} & \multicolumn{1}{c|}{\multirow{2}{*}{\begin{tabular}[c]{@{}c@{}}Inc. \\ Params (M)\end{tabular}}} & \multicolumn{1}{c|}{\multirow{2}{*}{base10}} & \multirow{2}{*}{de} & \multirow{2}{*}{pl} & \multirow{2}{*}{cy} & \multirow{2}{*}{ja} & \multicolumn{1}{c|}{\multirow{2}{*}{cs}} & \multirow{2}{*}{Avg} \\
\multicolumn{2}{c|}{}                         & \multicolumn{1}{c|}{}                                                                        & \multicolumn{1}{c|}{}                        &                     &                     &                     &                     & \multicolumn{1}{c|}{}                    &                      \\ \midrule
\multicolumn{2}{c|}{Mono}                     & \multicolumn{1}{c|}{45.5}                                                                    & \multicolumn{1}{c|}{21.1}                    & 15.1                & 10.8                & 15.5                & 19.3                & \multicolumn{1}{c|}{83.8}                & 22.2                 \\
\multicolumn{2}{c|}{Raw}                      & \multicolumn{1}{c|}{-}                                                                       & \multicolumn{1}{c|}{16.3}                    & 90.8                & 104.0               & 99.7                & 104.2               & \multicolumn{1}{c|}{103.1}               & 41.6                 \\
\multicolumn{2}{c|}{FullFT}                   & \multicolumn{1}{c|}{186.5}                                                                   & \multicolumn{1}{c|}{99.8}                    & \textbf{12.8}                & \textbf{8.0}                 & \textbf{10.5}                & \textbf{8.5}                 & \multicolumn{1}{c|}{\textbf{24.5}}                & 66.4                 \\
\multicolumn{2}{c|}{ER (1k)}                  & \multicolumn{1}{c|}{-}                                                                       & \multicolumn{1}{c|}{48.3}                    & 14.9                & 11.6                & 13.3                & 10.5                & \multicolumn{1}{c|}{29.5}                & 35.1                 \\
\multicolumn{2}{c|}{ER (10k)}                 & \multicolumn{1}{c|}{-}                                                                       & \multicolumn{1}{c|}{23.4}                    & 15.2                & 14.0                & 14.2                & 11.1                & \multicolumn{1}{c|}{30.9}                & 20.0                 \\
\multicolumn{2}{c|}{CJT}                      & \multicolumn{1}{c|}{-}                                                                       & \multicolumn{1}{c|}{\textbf{14.8}}                    & 15.2                & 12.5                & 15.9                & 13.3                & \multicolumn{1}{c|}{32.0}                & \textbf{14.8}                 \\ \midrule
$\bm{\alpha}$                         & PEFT         &                                                                                              & \multicolumn{7}{c}{PELE framework}                                                                                                                                                                     \\ \midrule
LP post                        & Adapter      & \multicolumn{1}{c|}{11.2}                                                                    & \multicolumn{1}{c|}{16.3}                    & 23.0                & 24.4                & 17.1                & 15.1                & \multicolumn{1}{c|}{35.4}                & 17.4                 \\
LP ohot                        & Adapter      & \multicolumn{1}{c|}{11.2}                                                                    & \multicolumn{1}{c|}{16.3}                    & 24.1                & 27.9                & 16.9                & 14.6                & \multicolumn{1}{c|}{32.4}                & 17.4                 \\ \hline
\multirow{3}{*}{GT ohot}       & LoRA         & \multicolumn{1}{c|}{11.3}                                                                    & \multicolumn{1}{c|}{16.3}                    & 31.0                & 33.3                & 29.0                & 29.7                & \multicolumn{1}{c|}{49.2}                & 21.0                 \\
                               & LoRA*        & \multicolumn{1}{c|}{16.2}                                                                    & \multicolumn{1}{c|}{16.3}                    & 28.1                & 26.4                & 24.5                & 22.7                & \multicolumn{1}{c|}{43.1}                & 19.2                 \\
                               & Adapter      & \multicolumn{1}{c|}{11.2}                                                                    & \multicolumn{1}{c|}{16.3}                    & 19.6                & 14.2                & 13.9                & 12.7                & \multicolumn{1}{c|}{\textbf{26.5}}                & \textbf{15.6}                 \\ \hline
GT learn                       & Adapter      & \multicolumn{1}{c|}{11.2}                                                                    & \multicolumn{1}{c|}{16.3}                    & \textbf{18.3}                & \textbf{12.8}                & \textbf{13.9}                & \textbf{12.4}                & \multicolumn{1}{c|}{29.9}                & 15.7                 \\ \bottomrule
\end{tabular}}
\label{tab:res}
\vspace{-15pt}
\end{table}

\subsection{Comparison between PELE and Baselines}
Tab. \ref{tab:res} presents the performance comparison of different methods on the MASR language extension. Five never-seen languages, namely German (de), Polish (pl), Welsh (cy), Japanese (ja), and Czech (cs), are utilized for MASR continual learning.  Compared to \textit{Raw}, \textit{FullFT} significantly increases the recognition performance on the new languages by fully fine-tuning. However, without access to the original languages' data, the CF issue is obvious. In this way, to extend a new language each time, a copy of the base model is required. In experience reply (ER), \textit{ER (10k)} alleviates the CF phenomenon more than \textit{ER (1k)} due to the relatively larger cached data size of the historical data. With no access limitation to previous languages' data, \textit{CJT} can naturally solve the catastrophic forgetting. It can be observed that \textit{CJT} further improves the recognition of the original 10 languages by continual training, while the performance on the new languages degrades compared to \textit{FullFT}. 

In the PELE block, \textit{Adapter}, as the best PEFT candidate shown in Sec. \ref{subsubsec: XLA}, are mainly experimented with different $\bm{\alpha}$ settings. The WER performance in the original 10 languages remains $16.3\%$, the same as \textit{Raw} for all the PELE configurations. Note that previous languages' data are not required in PELE. The CF issue is fundamentally solved due to the architecture-based CL design. Regarding LP, MLP\_6 is utilized to produce LID posterior or onehot. 
The overall average recognition performance only left behind \textit{CJT}. To observe the performance upper bound, $GT~ohot$ is introduced to exclude the language prediction (LP) error effect. In addition, \textit{LoRA} and \textit{LoRA*}, as the representative type of \textit{parameter composition} PEFT,  are also reported. \textit{Adapter} achieves at least $10\%$ absolute WER reduction per new language compared to \textit{LoRA}-related methods. The overall WER performance arrives at $15.6\%$, approaching $14.8\%$ of \textit{CJT}. Particularly, in three of five new languages, our best PELE system surpasses \textit{CJT}.  By comparing $LP~ohot$ with $GT~ohot$, we can see the performance degradation caused by the LP error. The frozen encoder layers make the MLP\_6 accuracy limited, thus affecting the overall CL performance. Better LP, maybe an external predictor, is expected to have less degradation. Furthermore, we attempt $\bm{\alpha}$ a learnable weight coefficient vector ($GT~learn$), which is initialized as $GT~ohot$.  By explicitly leveraging the interaction of multilingual module-level information, most languages exhibit performance improvement except the language Czech (cs).   


\section{Conclusions}
In this paper, we present a theoretically inspired architecture-based framework, PELE, for language extension in MASR. 
PELE is designed to be parameter-efficient for the framework's scalability. Many different PEFT modules are explored as candidates to perform cross-lingual adaptation. Experiments carried out on 5  never-seen languages show the effectiveness and efficiency of PELE. The best-performing PEFT module configuration attains an overall satisfactory performance, surpassing the continual joint learning setting on three of five languages. Importantly, it is found that the performance of PEFT in language extension is limited when working on weight parameters or input features. The PELE framework is flexible and more configurations (e.g., other PEFTs/coefficient vectors, curriculum training scheme, and the order of extended languages, etc) are expected to be further explored in future works.

\bibliographystyle{IEEEtran}
\bibliography{mybib}

\end{document}